\documentclass[runningheads]{llncs}
\usepackage[T1]{fontenc}
\usepackage{graphicx}
\usepackage{amsfonts}
\usepackage{xcolor}
\usepackage{amsmath}
\usepackage{adjustbox}
\usepackage{ulem}
\usepackage{booktabs,makecell, multirow, tabularx}
\usepackage{comment}

\newcommand{\xc}[1]{\textcolor{red}{[{\bf XC}: #1]}}
\newcommand{\kf}[1]{\textcolor{cyan}{[{\bf Keqiang}: #1]}}

\begin{document}
\title{IIHT: Medical Report Generation with  Image-to-Indicator Hierarchical Transformer}
\titlerunning{Enhancing Radiology Reports with IIHT }
%
\author{Keqiang Fan
\inst{1}
\and
Xiaohao Cai
\inst{1}
\and
Mahesan Niranjan
\inst{1}
}
\authorrunning{K. Fan et al.}
%
\institute{School of Electronics and Computer Science, University of Southampton, Southampton SO17 1BJ, UK
\\
\email{\{k.fan, x.cai, mn\}@soton.ac.uk}
}
\maketitle              

\begin{abstract}
Automated medical report generation has become increasingly important in medical analysis. It can produce computer-aided  diagnosis descriptions and thus significantly alleviate the doctors' work. 
Inspired by the huge success of neural machine translation and image captioning, various deep learning methods have been proposed for medical report generation. 
However, due to the inherent properties of medical data, including data imbalance and the length and correlation between report sequences, the generated  reports  by existing methods may exhibit linguistic fluency but lack adequate clinical accuracy.
In this work, we propose an image-to-indicator hierarchical transformer (IIHT) framework for medical report generation. It consists of three modules, i.e., a classifier module, an indicator expansion module and a generator module. 
The classifier module first extracts image features from the input medical images and produces disease-related indicators with their corresponding states. The disease-related indicators are subsequently utilised as input for the indicator expansion module, incorporating the "data-text-data" strategy. The transformer-based generator then leverages these extracted features along with image features as auxiliary information to generate final reports.
Furthermore, the proposed IIHT method is feasible for radiologists to modify disease indicators in real-world scenarios and integrate the operations into the indicator expansion module for fluent and accurate medical report generation. Extensive experiments and comparisons with state-of-the-art methods under various evaluation metrics demonstrate the great performance of the proposed method. 


\end{abstract}
\section{Introduction}

Medical images (e.g. radiology and pathology images) and the corresponding reports serve as critical catalysts for disease diagnosis and treatment \cite{european2015medical}. 
A medical report generally includes multiple sentences describing a patient’s history symptoms and normal/abnormal findings from different regions within the medical images. However, in clinical practice, writing standard medical reports is tedious and time-consuming for experienced medical doctors and error-prone for inexperienced doctors. 
This is because the comprehensive analysis of e.g. X-Ray images necessitates a detailed interpretation of visible information, including the airway, lung, cardiovascular system and disability. 
Such interpretation requires the utilisation of foundational physiological knowledge alongside a profound understanding of the correlation with ancillary diagnostic findings, such as laboratory results, electrocardiograms and respiratory function tests. 
Therefore, the automatic report generation technology, which can alleviate the medics' workload  and effectively notify inexperienced radiologists regarding the presence of abnormalities, has garnered dramatic interest in both artificial intelligence and clinical medicine.

Medical report generation has a close relationship with image captioning \cite{huang2019attention,you2016image}.
The encoder-decoder framework is quite popular in image captioning, e.g., a CNN-based image encoder to extract the visual information and an RNN/LSTM-based report decoder to generate the textual information with visual attention \cite{jing2017automatic,yin2019automatic,li2018hybrid,wu-etal-2022-deltanet}.
With the recent progress in natural language processing, investigating transformer-based models as alternative decoders has been a growing trend for report generation \cite{wang2022prior,chen2022cross,chen-emnlp-2020-r2gen,nguyen2022eddie}. 
The self-attention mechanism employed inside  the transformer can effectively eliminate information loss, thereby maximising the preservation of visual and textual information in the process of generating medical reports.
Although these methods have achieved remarkable performance and can obtain language fluency reports, limited studies have been dedicated to comprehending the intrinsic medical and clinical problems. The first problem is \textit{data imbalance}, e.g., the normal images dominate the dataset over the abnormal ones \cite{shin2016learning} and, for the abnormal images, normal regions could encompass a larger spatial extent than abnormal regions \cite{liu2022competence}. The narrow data distribution could make the descriptions of normal regions dominate the entire report. On the whole, imbalanced data may degrade the quality of the automatically generated reports, or even result in all generated reports being basically similar.
The second problem is \textit{length and correlation between report sequences}. Medical report generation is designed to describe and record the patient's symptoms from e.g. radiology images including cardiomegaly, lung opacity and fractures, etc. 
The description includes various disease-related symptoms and related topics rather than the prominent visual contents and related associations within the images, resulting in the correlation inside the report sequences not being as strong as initially presumed.
The mere combination of encoders (e.g. CNNs) and decoders (e.g. RNN, LSTM, and transformers) is insufficient to effectively tackle the aforementioned issues in the context of medical images and reports since these modalities represent distinct data types.
The above challenges motivate us to develop a more comprehensive method to balance visual  and textual features in unbalanced data
for medical report generation.

\begin{figure}
\vspace{-0.2cm}
\label{fig:framework}
\centering
\includegraphics[width=11.5cm]{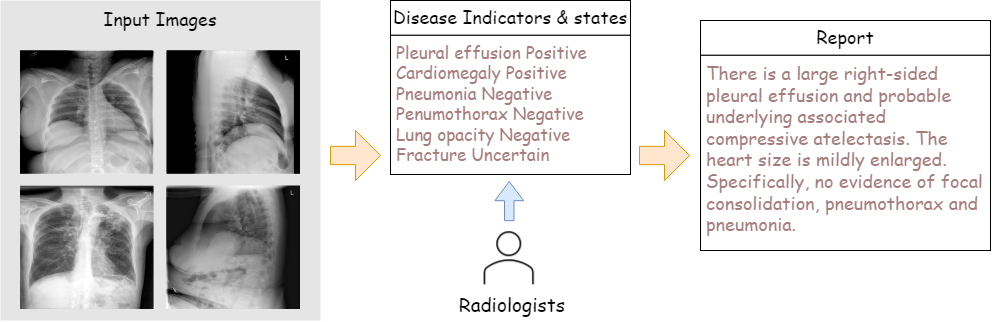}
\caption{The medical report writing procedure undertaken by radiologists.}
\vspace{-0.5cm}
\end{figure}

The radiologists’ working pattern in medical report writing is shown in Figure \ref{fig:framework}. Given a radiology image, radiologists first attempt to find the abnormal regions and evaluate the states for each disease indicator, such as uncertain, negative and positive. Then a correct clinical report is written through the stages for different indicators based on their working experience and prior medical knowledge. 
In this paper, we propose an image-to-indicator hierarchical transformer (IIHT) framework, imitating the radiologists’ working patterns (see Figure \ref{fig:framework}) to alleviate the above-mentioned problems in medical report generation. 


Our IIHT framework models the above working patterns through three modules: \textit{classifier}, \textit{indicator expansion} and \textit{generator}. The classifier module is an image diagnosis module, which could learn visual features and extract the corresponding disease indicator embedding from the input image. The indicator expansion module conducts the data-to-text progress, i.e., transferring the disease indicator embedding into short text sequences. The problem of data imbalance could be alleviated by encoding the indicator information, which models the domain-specific prior knowledge structure and summarises the disease indicator information and thus mitigates the long-sequence effects. Finally, the generator module produces the reports based on the encoded indicator information and image features. The whole generation pipeline is given in Figure \ref{figure:1}, which will be described in detail in Section \ref{sec:method}.
We remark that the disease indicator information here can also be modified by radiologists to standardise report fluency and accuracy. 
Overall, the contributions of this paper are three-fold:
\begin{itemize}
	\item[$\bullet$] We propose the IIHT framework, aiming to alleviate the data bias/imbalance problem and enhance the information correlation in long report sequences for medical report generation.
	\item[$\bullet$] We develop a dynamic approach which leverages integrated indicator information and allows radiologists to further adjust the report fluency and accuracy.
	\item[$\bullet$] We conduct comprehensive experiments and comparisons with state-of-the-art methods on the IU X-Ray dataset and demonstrate that our proposed method can achieve more accurate radiology reports.
\end{itemize}

The rest of the paper is organised as follows. Section \ref{sec:2} briefly recalls the related work in medical report generation. Our proposed method is introduced in Section \ref{sec:method}. Sections \ref{sec:4} and \ref{sec:5} present the details of the experimental setting and corresponding results, respectively. We conclude in Section \ref{sec:6}.

\section{Related Work}
\label{sec:2}

\noindent \textbf{Image captioning.}  The image captioning methods mainly adopt the encoder-decoder framework together with attention mechanisms \cite{you2016image} to translate the image into a single short descriptive sentence and have achieved great performance \cite{liang2017recurrent,vinyals2015show,anderson2018bottom,liu2018simnet}. Specifically, the encoder network extracts the visual representation from the input images and the decoder network generates the corresponding descriptive sentences. 
The attention mechanism enhances the co-expression of the visual features derived from the intermediate layers of CNNs and the semantic features from captions \cite{you2016image}. 
Recently, inspired by the capacity of parallel training, transformers \cite{vaswani2017attention} have been successfully applied to predict words according to multi-head self-attention mechanisms. However, these models demonstrate comparatively inferior performance on medical datasets as opposed to natural image datasets, primarily due to the disparity between homogeneous objects observed in different domains.
For instance, in the context of X-Ray images, there exists a relatively minimal discernible distinction between normal and abnormal instances, thereby contributing to the challenge encountered by models in accurately generating such captions.

\noindent \textbf{Medical report generation.}
Similar to image captioning, most existing medical report generation methods attempt to adopt a CNN-LSTM-based model to automatically generate fluent reports \cite{jing2017automatic,li2018hybrid,wang2018tienet,najdenkoska2021variational}. 
Direct utilisation of caption models often leads to the generation of duplicate and irrelevant reports. 
The work in \cite{jing2017automatic} developed a hierarchical LSTM model and a co-attention mechanism to extract the visual information and generate the corresponding descriptions. 
Najdenkoska et al. \cite{najdenkoska2021variational} explored variational topic inference to guide sentence generation by aligning image and language modalities in a latent space.
A two-level LSTM structure was also applied with a graph convolution network based on the knowledge graph to mine and represent the associations among medical findings during report generation \cite{wang2022prior}.
These methodologies encompass the selection of the most probable diseases or latent topic variables based on the sentence sequence or visual features within the data in order to facilitate sentence generation.
Recently, inspired by the capacity of parallel training, transformers \cite{zhou2018end,li2019entangled}
have successfully been applied to predict words according to the extracted features from CNN.
Chen et al. \cite{chen2022cross} proposed a transformer-based cross-modal memory network using a relational memory to facilitate interaction and generation across data modalities.
Nguyen et al. \cite{nguyen2022eddie} designed a differentiable end-to-end network to learn the disease feature representation and disease state to assist report generation.

The existing methods mentioned above  prioritise the enhancement of feature alignment between visual regions and disease labels. However, due to the inherent data biases and scarcity in the medical field, these models exhibit a bias towards generating reports that are plausible yet lack explicit abnormal descriptions.
Generating a radiology report is very challenging as it requires the contents of key medical findings and abnormalities with detailed descriptions for different data modalities. 
In this study, we address the challenges associated with data bias and scarcity in clinical reports through the utilisation of disease indicators as a bridge for more comprehensive medical report generation.

\section{Method}
\label{sec:method}

\begin{figure}[h]
\vspace{-0.5cm}
\includegraphics[width=12.3cm]{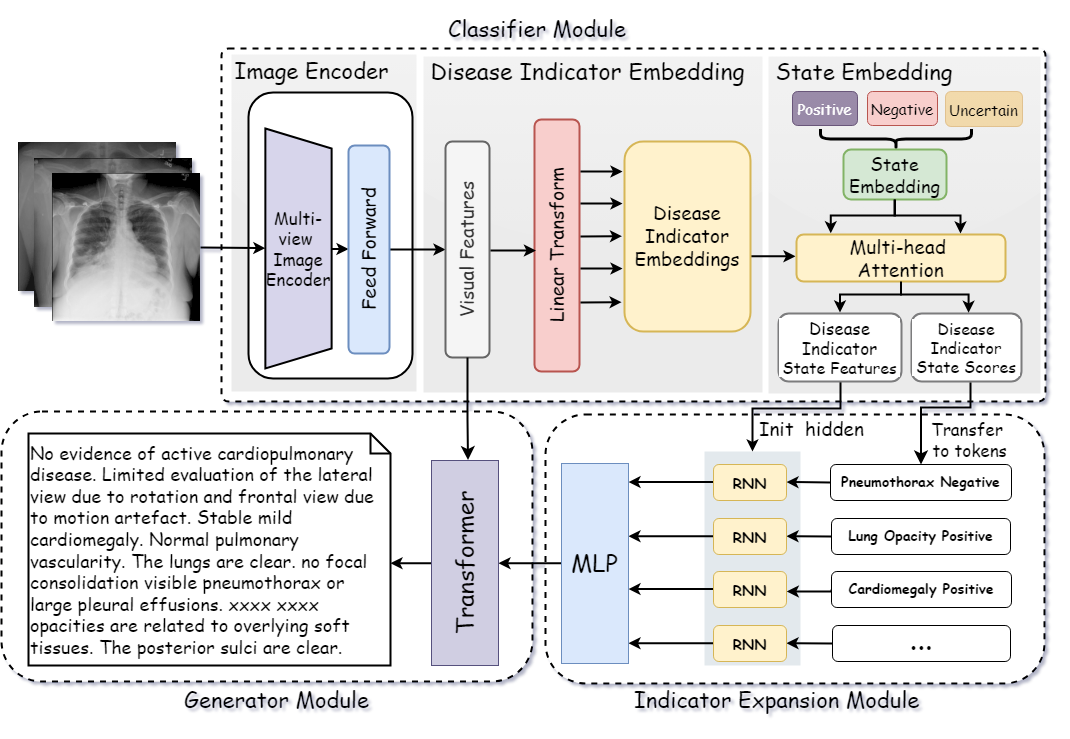}
\caption{The proposed IIHT framework. It consists of three modules: {classifier}, {indicator expansion} and {generator}. }
\label{figure:1}
\vspace{-0.5cm}
\end{figure}

An overview of our proposed IIHT framework is demonstrated in Figure \ref{figure:1}.
It follows the distinct stages involved in generating a comprehensive medical imaging diagnosis report, adhering to the established process employed in clinical radiology (e.g. see Figure \ref{fig:framework}). 

Given a radiology image $\textbf{I}$, the corresponding different indicators  are all classified into different states (e.g. positive, negative, uncertain, etc.) denoted as $\textbf{C} = \{\textbf{c}_1,\cdots,\textbf{c}_t,\cdots,\textbf{c}_T\}$, where $T$ is the number of indicators and $\textbf{c}_t$ is the one-hot encoding of the states. Particularly, these indicators can also be modified by radiologists to standardise the disease states across patients, thereby enhancing the correctness of the final generated report.
The corresponding generated report for a given radiology image is denoted as $\textbf{y} = (y_1,\cdots,y_n,\cdots,y_N)$, where $y_n \in \mathbb{V}$ is the generated unigram tokens, $N$ is the length of the report, and $\mathbb{V}$ is the vocabulary of all possible $v$ tokens for reports generation. For example, the word sequence "Pleural effusion" is segmented into small pieces of tokens, i.e., \{"Pleural", "effus", "ion"\}.
Generally, the aim of the report generation is to maximise the conditional log-likelihood, i.e.,
\begin{equation}
\theta^*=  \underset{\theta}{\arg \max } \prod_{n=1}^N p_{\theta}\left(y_n \mid y_1, \ldots, y_{n-1}, \mathbf{I} \right),
\end{equation} 
where $\theta$ denotes the model parameters and $y_0$ represents the start token.
After incorporating each disease indicator $\textbf{c} \in \textbf{C}$ into the conditional probability $ p_{\theta}\left(y_n \mid y_1, \ldots, y_{n-1}, \mathbf{I} \right)$, we have
\begin{equation}
\log p_{\theta}\left(y_n \mid y_1, \ldots, y_{n-1}, \mathbf{I} \right)=\int_{\textbf{C}} \log p_{\theta}\left(y_n \mid y_1, \ldots, y_{n-1}, \mathbf{c},  \mathbf{I} \right) p_\theta(\mathbf{c} \mid \mathbf{I}) {\rm d} \mathbf{c},
\end{equation} where $p_\theta(\mathbf{c}\mid\mathbf{I})$ represents the classifier module.

Recall that our IIHT framework is demonstrated in Figure \ref{figure:1}. The details are described in the subsections below.

\subsection{Classifier Module}
\noindent \textbf{Image encoder.} The first step in medical report generation is to extract the visual features from the given medical images.
In our research, we employ a pre-trained visual feature extractor, such as ResNet \cite{he2016deep}, to extract the visual features from patients' radiology images that commonly contain multiple view images. 
For simplicity, given a set of $r$  radiology images $\left\{\textbf{I}_{i}\right\}_{i=1}^{r}$, the final visual features say $\textbf{x}$ are obtained by merging the corresponding features of each image using max-pooling across the last convolutional layer.  The process is formulated as $\textbf{x}=f_v\left(\textbf{I}_{1}, \textbf{I}_{2}, \cdots,\textbf{I}_{r}\right)$, where $f_v\left(\cdot \right)$ refers to the visual extractor and $\textbf{x} \in \mathbb{R}^{F}$ with $F$ number of features.

\noindent \textbf{Capture disease indicator embedding.}
The visual features are further transformed into multiple low-dimensional feature vectors, regarded as disease indicator embeddings, which have the capacity to capture interrelationships and correlations among different diseases. The indicator disease embedding is denoted as $\mathbf{D} = (\textbf{d}_1, \cdots, \textbf{d}_T) \in \mathbb{R}^{e \times T}$, where $e$ is the embedding dimension and note that $T$ is the number of indicators. 
Each vector $\textbf{d}_t \in \mathbb{R}^{e}, t = 1, \cdots, T$ is the representation of the corresponding disease indicator, which can be acquired through a linear transformation of the visual features, i.e.,
\begin{equation}
    \textbf{d}_{t}=\textbf{W}_{t}^{\top} \textbf{x}+\textbf{b}_{t},
\end{equation}
where $\textbf{W}_{t} \in \mathbb{R}^{F \times e}$ and $\textbf{b}_{t} \in \mathbb{R}^{e}$ are learnable parameters of the $t$-th disease representation.

The intuitive advantage of separating high-dimensional image features into distinct low-dimensional embeddings is that it facilitates the exploration of the relationships among disease indicators.
However, when dealing with medical images, relying solely on disease indicator embeddings is insufficient due to the heterogeneous information, including the disease type (e.g. disease name) and the disease status (e.g. positive or negative). 
Consequently, we undertake further decomposition of the disease indicator embedding, thereby leading to the conception of the subsequent state embedding.

\noindent \textbf{Capture state embedding.}
To improve the interpretability of the disease indicator embeddings, a self-attention module is employed to offer valuable insights into the representation of each indicator.  
Each indicator embedding is further decomposed to obtain the disease state such as positive, negative or uncertain. 
Let $M$ be the number of
states and  $\textbf{S} = (\textbf{s}_1, \cdots, \textbf{s}_M) \in \mathbb{R}^{e \times M}$ be the state embedding, which is randomly initialized and learnable.
Given a disease indicator embedding vector $\textbf{d}_{t}$, the final state-aware of the disease embedding say $\hat{\textbf{d}}_{t} \in \mathbb{R}^{e}$ is obtained by 
$\hat{\textbf{d}}_{t} =\sum_{m=1}^{M} \alpha_{t m} \textbf{s}_{m}$, where $\alpha_{tm}$ is the self-attention score of $\textbf{d}_{t}$ and $\textbf{s}_{m}$ defined as
\begin{equation}
    \alpha_{t m}=\frac{\exp ({\textbf{d}_{t}^{\top} \cdot \textbf{s}_{m}})}{\sum_{m=1}^{M} \exp{(\textbf{d}_{t}^{\top} \cdot \textbf{s}_{m})}}.
\end{equation}
Iteratively, each disease indicator representation $\textbf{d}_t$ will be matched with its corresponding state embedding $\textbf{s}_m$ by computing vector similarity, resulting in an improved disease indicator representation $\hat{\textbf{d}}_{t}$.

\noindent \textbf{Classification.}
To enhance the similarity between $\textbf{d}_t$ and $\textbf{s}_m$, we treat this as a multi-label problem. The calculated self-attention score $ \alpha_{t m}$ is the confidence level of classifying disease $t$ into the state $m$, which is then used as a predictive value. 
By abuse of notation, let $\textbf{c}_t = \{c_{t1},\cdots,c_{tm},\cdots,c_{tM} \}$ be the $t$-th ground-true disease indicator and 
$\boldsymbol{\alpha}_t = \{\alpha_{t1},\cdots,\alpha_{tm},\cdots,\alpha_{tM} \}$ be the prediction, where $c_{tm} \in \{0, 1\}$ and $\alpha_{tm} \in \left(0,1\right)$.
The loss of the multi-label classification can be defined as
\begin{equation}
    \mathcal{L}_{C}=-\frac{1}{T} \sum_{t=1}^{T} \sum_{m=1}^{M} c_{t m} \log \left(\alpha_{t m}\right).
\end{equation} 
The maximum value $\alpha_{t m}$ in $\boldsymbol{\alpha}_t$ represents the predicted state for disease $t$.
To enable integration with the indicator expansion module, we adopt an alternative approach; instead of directly utilizing $\hat{\textbf{d}}_{t}$, we recalculate the state-aware embedding for the $t$-th disease indicator, denoted as $\hat{\textbf{s}}_{t} \in \mathbb{R}^e$, i.e.,
\begin{equation}
    \hat{\textbf{s}}_{t}= \sum_{m=1}^{M} \left\{\begin{array}{lc}
 c_{tm} \textbf{s}_m, & \text {if training phase}, \\
\alpha_{tm} \textbf{s}_m, & \text {otherwise}.
\end{array}\right.
\end{equation}
Hence, the state-aware disease indicator embedding $ \hat{\textbf{s}}_{t}$ directly contains the state information of the disease $t$.

\subsection{Indicator Expansion Module}
In the indicator extension module, we employ a "data-text-data" conversion strategy. This strategy involves converting the input indicator embedding from its original format into a textual sequential word representation and then converting it back to the original format.
The inherent interpretability of short disease indicator sequences can be further enhanced, resulting in generating more reliable medical reports. 
For each disease indicator and its state, whether it is the ground-truth label $ \textbf{c}_{t}$ or the predicted label $\boldsymbol{\alpha}_{t}$, it can be converted into a sequence of words, denoted as $\hat{\textbf{c}}_{t} =\{\hat{c}_{t1},\cdots,\hat{c}_{tk},\cdots,\hat{c}_{tK}\}$, where $\hat{c}_{tk} \in \mathbb{W}$ is the corresponding word in the sequence, $K$ is the length of the word sequence, and $\mathbb{W}$ is the vocabulary of all possible words in all indicators. For example, an indicator such as "lung oedema uncertain" can be converted into a word sequence such as \{"lung", "oedema", "uncertain"\}. To extract the textual information within the short word sequence for each disease $t$, we use a one-layer bi-directional gated recurrent unit  as an encoder say $f_{w}\left(\cdot \right)$ followed by a multi-layer perceptron (MLP) $\mathbf{\Phi}$ to generate  the indicator information $\textbf{h}_{t} \in \mathbb{R}^{e}$, i.e., 
\begin{equation}
 \textbf{h}_{t} 
 =\mathbf{\Phi} \left(\textbf{h}_{t 0}^{w} + \textbf{h}_{t k}^{w}\right), \ \ 
    \textbf{h}_{t k}^{w}  =f_{w} \left(\hat{c}_{t k}, \textbf{h}_{t k-1}^{w}\right), 
\end{equation} 
where $\textbf{h}_{t k}^{w} \in \mathbb{R}^{e}$ is the hidden state in $f_{w}$. 
For each disease indicator, the initial state $\left(k=0\right)$ in $f_{w}$ is the corresponding state-aware disease indicator embedding $\hat{\textbf{s}}_{t}$, i.e., $\textbf{h}_{t 0}^{w}=\hat{\textbf{s}}_{t}$.

\subsection{Generator Module}
The {generator} say $f_g$ of our IIHT framework is based on the transformer encoder architecture, comprising $Z$ stacked masked multi-head self-attention layers alongside a feed-forward layer positioned at the top of each layer. 
Each word $y_{k}$ in the ground-truth report is transferred into the corresponding word embedding $\hat{\textbf{y}}_{k} \in \mathbb{R}^{e}$. 
For the new word $y_{n}$, the hidden state representation $\textbf{h}_{n}^{\prime} \in \mathbb{R}^{e}$ in the generator $f_g$ is computed based on the previous word embeddings $\left\{\hat{\textbf{y}}_{k}\right\}_{k=1}^{n-1}$, the calculated indicator information $\left\{\textbf{h}_{t}\right\}_{t=1}^{T}$ and the visual representation $\textbf{x}$, i.e.,
\begin{equation}
    \textbf{h}^{\prime}_n =  f_{g}\left(\hat{\textbf{y}}_1, \ldots, \hat{\textbf{y}}_{n-1},\textbf{h}_{1},\cdots, \textbf{h}_{T}, \textbf{x} \right).
\end{equation}
For the $i$-th report, the confidence $\textbf{p}_{n}^{i} \in \mathbb{R}^{v}$ of the word $y_n$ is calculated by
\begin{equation}
    \textbf{p}_{n}^{i}=\operatorname{softmax}
    \left(\textbf{W}_p^{\top} \textbf{h}^{\prime}_n\right),
\end{equation} where $\textbf{W}_{p} \in \mathbb{R}^{e\times v}$ is a learnable parameter and recall that $v$ is the size of $\mathbb{V}$. 

The loss function of the generator say $\mathcal{L}_{\mathcal{G}}$ is determined based on the cross-entropy loss, quantifying all the predicted words in all the given $l$ medical reports with their ground truth, i.e.,
\begin{equation}
\mathcal{L}_{\mathcal{G}}=-\frac{1}{l} \sum_{i=1}^{l} \sum_{n=1}^{N} \sum_{j=1}^{v} y_{nj}^{i} \log \left(p_{n j}^{i}\right),
\end{equation} 
where $p_{n j}^{i}$ is the $j$-th component of $\textbf{p}_{n}^{i}$, and $y_{nj}^{i} $ is $j$-th component of $\textbf{y}_{n}^{i} \in \mathbb{R}^{v}$ which is the ground-truth one-hot encoding for word $y_n$ in the $i$-th report.
Therefore, the final loss of our IIHT method is 
\begin{equation}
    \mathcal{L}=\lambda \mathcal{L}_{\mathcal{G}} +\left(1-\lambda \right) \mathcal{L}_{C},
    \label{total_loss}
\end{equation} where $\lambda$ is a hyperparameter.

\section{Experimental Setup}
\label{sec:4}

\subsection{Data}

The publicly available IU X-Ray dataset \cite{demner2016preparing} is adopted for our evaluation. It contains 7,470 chest X-Ray images associated with 3,955 fully de-identified medical reports. 
Within our study, each report comprises multi-view chest X-Ray images along with distinct sections dedicated to impressions, findings and indications.

\subsection{Implementation}
Our analysis primarily focuses on reports with a finding section, as it is deemed a crucial component of the report. 
To tackle the issue of data imbalance, we utilise a strategy wherein we extract 11 prevalent disease indicators from the dataset, excluding the "normal" indicators based on the findings and indication sections of the reports.
Additionally, three states (i.e., {uncertain}, {negative} and {positive}) are assigned to each indicator. 
In cases where a report lacks information regarding all indicators, we discard the report to ensure data integrity and reliability. 
The preprocessing of all reports is followed by the random selection of image-report pairs, which are then divided into three sets, i.e., training, validation and test sets. The distribution of these sets is 70\%, 10\% and 20\%, respectively.
All the words in the reports are segmented into small pieces by SentencePiece \cite{kudo-richardson-2018-sentencepiece}.
Standard five-fold cross-validation on the training set is used for model selection.


To extract visual features, we utilise two different models: ResNet-50 \cite{he2016deep} pre-trained on ImageNet \cite{deng2009imagenet} and a vision transformer (ViT) \cite{dosovitskiy2020image}. 
Prior to extraction, the images are randomly cropped to a size of $224 \times 224$, accompanied by data augmentation techniques.
Within our model, the disease indicator embedding, indicator expansion module and generator module all have a hidden dimension of $512$. 
During training, we iterate 300 epochs with a batch size of $8$. The hyperparameter $\lambda$ in the loss function is set to $0.5$. For optimisation, we employ AdamW \cite{loshchilov2017decoupled} with a learning rate of $10^{-6}$ and a weight decay of $10^{-4}$.

\subsection{Metrics}
The fundamental evaluation concept of the generated reports is to quantify the correlation between the generated and the ground-truth reports.  Following most of the image captioning methods, we apply the most popular metrics for evaluating natural language generation such as 1--4 gram BLEU \cite{papineni2002bleu}, Rouge-L \cite{lin2004rouge} and METEOR \cite{denkowski2014meteor} to evaluate our model.

\section{Experimental Results}
\label{sec:5}
In this section, we first evaluate and compare our IIHT method with the state-of-the-art medical report generation methods. Then we conduct an ablation study for our method to verify the effectiveness of the indicator expansion module under different image extractors.

\begin{table}[htbp]
\begin{tabular}{clcccccc}
\hline
\multicolumn{2}{c}{Methods}            & BLEU-1         & BLEU-2         & BLEU-3         & BLEU-4         & METEOR         & ROUGE-L        \\ \hline \hline
\multicolumn{2}{c}{VTI \cite{najdenkoska2021variational}$^{\dagger}$}                & 0.493          & 0.360          & 0.291          & 0.154          & 0.218          & 0.375          \\
\multicolumn{2}{c}{Wang et al. \cite{wang2022prior}$^{\dagger}$}              & 0.450          & 0.301          & 0.213          & 0.158          & -              & 0.384         \\
\multicolumn{2}{c}{CMR \cite{chen2022cross}$^{\dagger}$}     & 0.475          & 0.309          & 0.222          & 0.170          & 0.191          & 0.375          \\
\multicolumn{2}{c}{R2Gen \cite{chen-emnlp-2020-r2gen}$^{\dagger}$}   & 0.470          & 0.304          & 0.219          & 0.165          & 0.187          & 0.371          \\
\multicolumn{2}{c}{Eddie-Transformer \cite{nguyen2022eddie}$^{\dagger}$}   & 0.466          & 0.307          & 0.218          & 0.158          & -              & 0.358          \\
\multicolumn{2}{c}{CMAS \cite{jing-etal-2019-show}$^{\dagger}$} & 0.464          & 0.301          & 0.210          & 0.154          & -          & 0.362          \\
\multicolumn{2}{c}{DeltaNet \cite{wu-etal-2022-deltanet}$^{\dagger}$}                & 
0.485 & 0.324         & 0.238         &  0.184          & -          & 0.379         \\ \hline
\multicolumn{2}{c}{\textbf{Ours}}      
& \begin{tabular}[c]{@{}c@{}}\textbf{0.513} \\ $\pm$ \\ 0.006  \end{tabular}  
& \begin{tabular}[c]{@{}c@{}}\textbf{0.375} \\ $\pm$ \\ 0.005  \end{tabular} 
& \begin{tabular}[c]{@{}c@{}}\textbf{0.297} \\ $\pm$ \\ 0.006  \end{tabular} 
& \begin{tabular}[c]{@{}c@{}}\textbf{0.245} \\ $\pm$ \\ 0.006  \end{tabular} 
& \begin{tabular}[c]{@{}c@{}}\textbf{0.264} \\ $\pm$ \\ 0.002  \end{tabular} 
& \begin{tabular}[c]{@{}c@{}}\textbf{0.492} \\ $\pm$ \\ 0.004 \end{tabular} 
\\
\hline
\end{tabular}
\vspace{0.05cm}
\caption{Comparison between our IIHT method and the state-of-the-art medical report generation methods on the IU X-Ray dataset. Sign $\dagger$ refers to the results from the original papers. A higher value denotes better performance in all columns. }
\label{table:1}
\vspace{-0.5cm}
\end{table}

\subsection{ Report Generation}\label{sec:5.1}

We compare our method with the state-of-the-art medical report generation models, including the variational topic inference (VTI) framework \cite{najdenkoska2021variational}, a graph-based method to integrate prior knowledge in generation \cite{wang2022prior}, the cross-modal memory network (CMR) \cite{chen2022cross}, the memory-driven transformer (R2Gen) \cite{chen-emnlp-2020-r2gen}, the co-operative multi-agent system (CMAS) \cite{jing-etal-2019-show}, the enriched disease embedding based transformer (Eddie-Transformer) \cite{nguyen2022eddie}, and the conditional generation process for report generation (DeltaNet) \cite{wu-etal-2022-deltanet}. The quantitative results of all the methods on the IU X-Ray dataset are reported in Table \ref{table:1}. It clearly shows that our proposed IIHT method outperforms the state-of-the-art methods by a large margin across all the evaluation metrics, demonstrating the dramatic effectiveness of our method.

The methods under comparison in our study focus on exploring the correlation between medical images and medical reports. 
Some of these approaches have incorporated supplementary indicators as auxiliary information. 
However, these indicators primarily comprise frequently occurring phrases across all reports, disregarding the inherent imbalance within medical data. 
Consequently, the generated reports often treat abnormal patients as normal, since the phrases describing normal areas dominate the dataset.
 In contrast, our proposed method leverages  disease indicators  and assigns corresponding states based on the reported content. 
By adopting a "data-text-data" onversion approach in the indicator expansion module, our method effectively mitigates the issue of misleading the generated medical reports, and thus surpasses the performance of the existing approaches.

\subsection{Ablation Study}\label{sec:5.2}

\begin{table}[htbp]
\begin{tabular}{cl|c|cccccc}
\hline 
\multicolumn{2}{c|}{Methods}            & Encoder                    & BLEU-1 & BLEU-2 & BLEU-3 & BLEU-4 & METEOR & ROUGE-L \\ \hline \hline
\multicolumn{2}{c|}{\begin{tabular}[c]{@{}c@{}}IIHT\\ w/o Indicator\end{tabular} } &  \multirow{4}{*}{ViT}                          
& \begin{tabular}[c]{@{}c@{}}0.434 \\ $\pm$ \\ 0.002  \end{tabular}
& \begin{tabular}[c]{@{}c@{}}0.294 \\ $\pm$ \\ 0.004  \end{tabular}  
& \begin{tabular}[c]{@{}c@{}}0.210 \\ $\pm$ \\ 0.004  \end{tabular}
& \begin{tabular}[c]{@{}c@{}}0.153 \\ $\pm$ \\ 0.004  \end{tabular}
& \begin{tabular}[c]{@{}c@{}}0.216 \\ $\pm$ \\ 0.001  \end{tabular}
& \begin{tabular}[c]{@{}c@{}}0.409 \\ $\pm$ \\ 0.005  \end{tabular}
\\ \cline{0-1} \cline{4-9}
\multicolumn{2}{c|}{\begin{tabular}[c]{@{}c@{}}IIHT\\(Proposed)\end{tabular} }
&                            
& \begin{tabular}[c]{@{}c@{}}0.463 \\ $\pm$ \\ 0.006  \end{tabular}
& \begin{tabular}[c]{@{}c@{}}0.323 \\ $\pm$ \\ 0.005  \end{tabular}
& \begin{tabular}[c]{@{}c@{}}0.241 \\ $\pm$ \\ 0.005  \end{tabular}
& \begin{tabular}[c]{@{}c@{}}0.186 \\ $\pm$ \\ 0.004  \end{tabular}
& \begin{tabular}[c]{@{}c@{}}0.234 \\ $\pm$ \\ 0.003  \end{tabular}
& \begin{tabular}[c]{@{}c@{}}0.445 \\ $\pm$ \\ 0.004  \end{tabular}
\\  \hline
\multicolumn{2}{c|}{\begin{tabular}[c]{@{}c@{}}IIHT\\ w/o Indicator\end{tabular} } & \multirow{4}{*}{ResNet-50}                       
& \begin{tabular}[c]{@{}c@{}}0.428  \\ $\pm$ \\ 0.007  \end{tabular} 
& \begin{tabular}[c]{@{}c@{}}0.271  \\ $\pm$ \\ 0.008  \end{tabular}
& \begin{tabular}[c]{@{}c@{}}0.188  \\ $\pm$ \\ 0.003  \end{tabular}
& \begin{tabular}[c]{@{}c@{}}0.136  \\ $\pm$ \\ 0.003  \end{tabular}
& \begin{tabular}[c]{@{}c@{}}0.185 \\ $\pm$ \\ 0.002  \end{tabular}
&  \begin{tabular}[c]{@{}c@{}}0.376 \\ $\pm$ \\ 0.004  \end{tabular}
\\ \cline{0-1} \cline{4-9}
\multicolumn{2}{c|}{\begin{tabular}[c]{@{}c@{}}IIHT\\(Proposed)\end{tabular} }
&                            
& \begin{tabular}[c]{@{}c@{}}\textbf{0.513} \\ $\pm$ \\ 0.006  \end{tabular}  
& \begin{tabular}[c]{@{}c@{}}\textbf{0.375} \\ $\pm$ \\ 0.005  \end{tabular} 
& \begin{tabular}[c]{@{}c@{}}\textbf{0.297} \\ $\pm$ \\ 0.006  \end{tabular} 
& \begin{tabular}[c]{@{}c@{}}\textbf{0.245} \\ $\pm$ \\ 0.006  \end{tabular} 
& \begin{tabular}[c]{@{}c@{}}\textbf{0.264} \\ $\pm$ \\ 0.002  \end{tabular} 
& \begin{tabular}[c]{@{}c@{}}\textbf{0.492} \\ $\pm$ \\ 0.004  \end{tabular} 
\\
\hline
\end{tabular}
\vspace{0.05cm}
\caption{
The ablation study of our method on the IU X-Ray dataset. "w/o Indicator" refers to the model without the indicator expansion module.
}
\label{table:2}
\vspace{-0.9cm}
\end{table}

\begin{table}[htbp]
\centering
\setlength{\tabcolsep}{0.2mm}{}
\begin{tabular}{|c|p{2.5cm}|p{2.5cm}|}
\hline
Data             & \multicolumn{1}{c|}{\begin{tabular}[c]{@{}l@{}}Groud-truth  Reports\end{tabular}}   & \multicolumn{1}{c|}{Generated Reports}     \\ \hline \hline
\multicolumn{1}{|c|}{
\begin{minipage}[c]{0.25\columnwidth}
\centering
\raisebox{-.52\height}{\includegraphics[width=0.7\linewidth,height=0.55\linewidth]{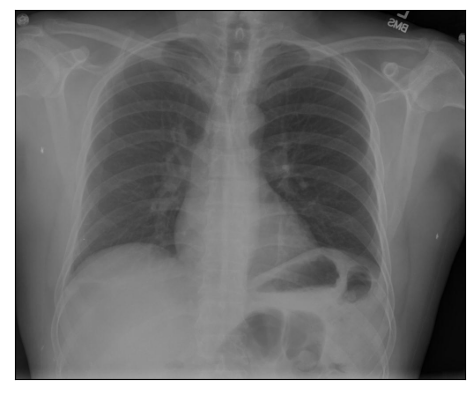}}
\raisebox{-.45\height}{\includegraphics[width=0.7\linewidth,height=0.55\linewidth]{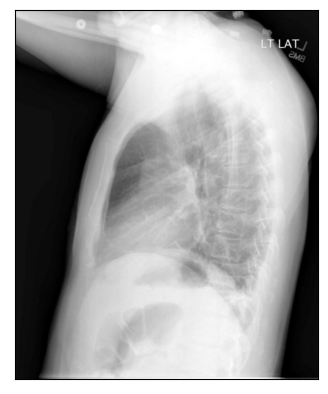}}
\end{minipage}  
}    
& \multicolumn{1}{l|}{
\begin{tabular}[c]{@{}p{4.4cm}@{}} No acute cardiopulmonary findings. No focal consolidation. No visualized pneumothorax. No large pleural effusions . The heart size and cardiomediastinal silhouette are grossly unremarkable. \end{tabular}
}  
& 
\multicolumn{1}{l|}{
\begin{tabular}[c]{@{}p{4.4cm}@{}}No acute cardiopulmonary abnormality. The lungs are clear bilaterally. Specifically, no evidence of focal consolidation pneumothorax or pleural effusion. Cardiomediastinal silhouette is unremarkable. visualized osseous structures of the thorax are without acute abnormality.\end{tabular}
}\\
\hline
\multicolumn{3}{|l|}{\begin{tabular}[c]{@{}p{12.3cm}@{}}\textbf{Indicators:} Cardiomediastinal silhouette negative; pneumothorax negative; granuloma negative; consolidation negative; pleural effusion negative; pneumonia negative.\end{tabular}}                           \\ \hline \hline
\multicolumn{1}{|l|}{
\begin{minipage}[c]{0.25\columnwidth}
\centering
\raisebox{-.45\height}{\includegraphics[width=0.7\linewidth,height=0.55\linewidth]{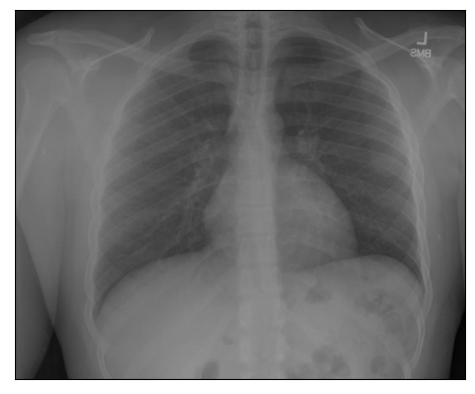}}
\raisebox{-.45\height}{\includegraphics[width=0.7\linewidth,height=0.55\linewidth]{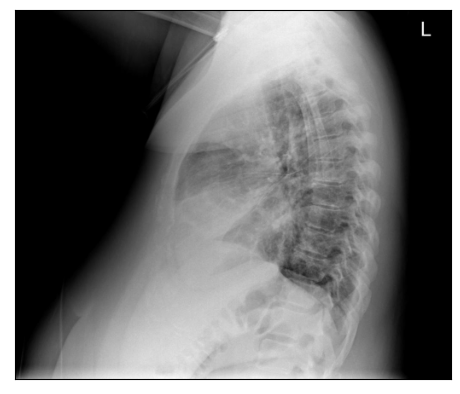}}
\end{minipage}  
}    
& \multicolumn{1}{l|}{
\begin{tabular}[c]{@{}p{4.4cm}@{}} Right middle lobe and lower lobe pneumonia. Heart size is within the upper limits of the normal. The pulmonary and mediastinum are within normal limits. there is no pleural effusion or pneumothorax. There is the right basilar air space opacity. \end{tabular}
}  
& 
\multicolumn{1}{l|}{
\begin{tabular}[c]{@{}p{4.4cm}@{}}Right lower lobe airspace disease in the right lower lobe atelectasis or pneumonia. Heart size and pulmonary vascularity appear within normal limits. There is no pleural effusion or pneumothorax. There are no acute bony abnormalities.\end{tabular}
}\\
\hline
\multicolumn{3}{|l|}{\begin{tabular}[c]{@{}p{12.3cm}@{}}\textbf{Indicators:} Lung opacity positive; pneumonia positive; pulmonary edema negative; pulmonary negative; pleural effusion negative; and pneumothorax negative.\end{tabular}}                           \\ \hline

\end{tabular}
\vspace{0.05cm}
\caption{Generated samples by our method on the IU X-Ray dataset. }
\label{table:3}
\end{table}


We now conduct an ablation study for our method to verify the effectiveness of different image extractors.
Table \ref{table:2} presents the results of our experiments, wherein we employed different visual feature extractors with and without the indicator expansion module. Specifically, we exclude the original "data-text-data" conversion strategy; instead, the disease indicator state features are directly used as the input of the MLP layer. This study allows us to analyse the influence of the "data-text-data" strategy within the indicator expansion module on the performance of the proposed IIHT framework.

By excluding the incremental disease indicator information, we observe that the image extractor ViT has a better performance than ResNet-50, see the results of the first and third rows in Table \ref{table:2}. This indicates that ViT is capable of effectively capturing semantic feature relationships within images. These findings provide evidence regarding the advantages of ViT in extracting visual information from images. 
We also observe that utilising indicator information extracted from the indicator  expansion module indeed contributes to the generation of precise and comprehensive medical reports, resulting in a noteworthy enhancement in terms of the quality of the generated reports. This improvement is observed when using both ViT and ResNet-50. 
Interestingly, as indicated in the second and fourth rows in Table \ref{table:2}, when the indicator expansion module is added, the performance improvement of ViT is not as significant as that of ResNet-50. We hypothesise that ViT requires a substantial amount of data to learn effectively from scratch. It is possible that the limited number of iterations during fine-tuning prevents ViT from achieving its full potential in performance enhancement. 
On the whole, our proposed IIHT method offers significant improvements over the state-of-the-art models. This enhancement can be attributed to the inclusion of the disease indicator expansion module, which plays a crucial role in enhancing the quality of the generated reports.

Finally, in Table \ref{table:3}, we showcase some examples of the reports generated by our method. By incorporating both images and indicators, our method closely mimics the process followed by radiologists when composing medical reports while also addressing the data imbalance challenge. 
Even in the case where all indicators are normal, a generated report for a healthy patient typically includes a description of various disease indicators, as shown in the first example in Table  \ref{table:3}. 
For patients with abnormal conditions, our method still has a remarkable ability to accurately generate comprehensive reports.
Moreover, our method incorporates the capability of facilitating real-time modification of disease indicators, thereby enabling a more accurate and complete process for report generation. This functionality serves to minimise the occurrence of misdiagnosis instances, and thus enhances the overall accuracy and reliability of the generated reports.
As a result, we reveal that the generated medical reports with the use of indicator-based features can be more reasonable and disease-focused in comparison to traditional "image-to-text" setups.

\section{Conclusion}
\label{sec:6}
In this paper, we proposed a novel method called IIHT for medical report generation by integrating disease indicator information into the report generation process. 
The IIHT framework consists of the {classifier} module, {indicator expansion} module and {generator} module. 
The "data-text-data" strategy implemented in the indicator expansion module leverages the textual information in the form of concise phrases extracted from the disease indicators and states. The accompanying data conversion step enhances the indicator information, effectively resolving the data imbalance problem prevalent in medical data. Furthermore, this conversion also facilitates the correspondence between the length and correlation of medical data texts with disease indicator information.
Our method makes it feasible for radiologists to modify the disease indicators in real-world scenarios and integrate the operations into the {indicator expansion} module, which ultimately contributes to the standardisation of report fluency and accuracy. Extensive experiments and comparisons with state-of-the-art methods demonstrated the great performance of the proposed method. 
One potential limitation of our experiments is related to the accessibility and accuracy of the disease indicator information. The presence and precision of such disease indicator information can affect the outcomes of our study.
Interesting future work could involve investigating and enhancing our method from a multi-modal perspective by incorporating additional patient information such as age, gender and height for medical report generation.


%
%
%
\bibliographystyle{splncs04}
\bibliography{mybibliography}
%





\end{document}